\documentclass[conference,onecolumn]{IEEEtran}
\IEEEoverridecommandlockouts
\usepackage{cite}
\usepackage{amsmath,amssymb,amsfonts}
\usepackage{algorithmic}
\usepackage{graphicx}
\usepackage{textcomp}
\usepackage{xcolor}
\usepackage{float} 
\usepackage{multirow}
\usepackage{booktabs}
\def\BibTeX{{\rm B\kern-.05em{\sc i\kern-.025em b}\kern-.08em
    T\kern-.1667em\lower.7ex\hbox{E}\kern-.125emX}}

\makeatletter
\def\@cite#1#2{[{#1\if@tempswa , #2\fi}]}
\makeatother

\begin{document}

\title{Federated Learning for Multi-Center Sepsis Early Prediction with Privacy-Preserving \\

}

\author{\IEEEauthorblockN{Xixi Tian}
\IEEEauthorblockA{\textit{College of Computer and Information Science} \\
\textit{Southwest University}\\
Chongqing, China \\
sumonpoem@email.swu.edu.cn}
\and
\IEEEauthorblockN{Di Wu}
\IEEEauthorblockA{\textit{College of Computer and Information Science}  \\
\textit{Southwest University}\\
Chongqing, China \\
wudi1986@swu.edu.cn}
\and
\IEEEauthorblockN{Xiang Liu}
\IEEEauthorblockA{\textit{Department of Anesthesiology}\\
\textit{Southwest Hospital}\\
Chongqing, China \\
liuxiang@tmmu.edu.cn}
\and
\IEEEauthorblockN{Yiziting Zhu}
\IEEEauthorblockA{\textit{Department of Anesthesiology}\\
\textit{Southwest Hospital}\\
Chongqing, China \\
zhuyizyzt@tmmu.edu.cn}
\and
\IEEEauthorblockN{Yujie Li}
\IEEEauthorblockA{\textit{Department of Anesthesiology}\\
\textit{Southwest Hospital}\\
Chongqing, China \\
lyj09c@tmmu.edu.cn}
\and
\IEEEauthorblockN{Xin Shu}
\IEEEauthorblockA{\textit{Department of Anesthesiology}\\
\textit{Southwest Hospital}\\
Chongqing, China \\
shustarry@tmmu.edu.cn}

\and
\IEEEauthorblockN{Bin Yi}
\IEEEauthorblockA{\textit{Department of Anesthesiology}\\
\textit{Southwest Hospital}\\
Chongqing, China \\
yibin1974@tmmu.edu.cn}

}

\maketitle

\begin{abstract}
Privacy-sensitive and distributed characteristics of multi-center medical data bring severe obstacles to centralized modeling for accurate early prediction of sepsis. Federated learning (FL) has attracted growing attention as a promising framework for collaborative model development, as it allows multiple institutions to jointly train predictive models without directly sharing or centralizing raw data. Nevertheless, its practical performance, robustness, and privacy-preserving benefits remain insufficiently evaluated using real-world clinical datasets. To bridge this gap, this study systematically examines the application of federated learning to multi-center sepsis prediction. The experimental dataset consists of 648 clinically screened samples collected from three tertiary hospitals in China, with rigorous inclusion and exclusion criteria. We establish a centralized training paradigm as the performance baseline, and then implement a horizontal federated learning framework for distributed collaborative modeling. Extensive experimental results demonstrate that the federated learning-based model achieves highly comparable prediction accuracy to the centralized counterpart, while fundamentally avoiding privacy leakage. Further privacy security analysis verifies that malicious attackers cannot reconstruct the original patient data from the transmitted model parameters, indicating strong resistance against data reconstruction attacks. This work not only validates the practicality and security of federated learning in clinical sepsis prediction, but also provides a reliable and feasible solution for privacy-preserving multi-center medical collaboration. 
\end{abstract}


\section{Introduction}
Sepsis is a life-threatening condition resulting from a dysregulated host response to infection. It can cause severe organ dysfunction and remains associated with high mortality worldwide, contributing substantially to in-hospital deaths among surgical patients\cite{b46,b1,b55,b45}. Timely and accurate early prediction of sepsis is crucial, as patient conditions can deteriorate rapidly within hours, and delayed intervention directly increases mortality risk.

Traditional sepsis prediction methods using machine learning (Random Forest, XGBoost, SVM, Naïve Bayes, logistic regression) perform well on single-center, small-scale curated datasets but face prominent limitations in multi-center clinical scenarios. First, these methods typically depend on the centralized collection of raw clinical data, which is often infeasible in multi-center scenarios. This is mainly due to stringent privacy regulations and the sensitive nature of patient information, as direct data sharing among hospitals is frequently restricted to prevent potential privacy breaches. Second, these centralized models lack cross-institution generalizability, as medical data from different hospitals exhibit inherent heterogeneity, leading to performance degradation when deployed across multiple institutions\cite{b2,b50,b44,b3}. 

Federated learning (FL) is a distributed machine learning framework that enables multiple institutions to collaboratively train models without transferring raw data outside local sites. It has therefore become a promising approach for overcoming the aforementioned challenges. By exchanging and aggregating model parameters rather than patient-level data, FL helps protect data privacy while making use of multi-center information to improve model generalizability\cite{b52,b4,b51,b5}. However, the effectiveness and reliability of FL in real-world multi-center sepsis prediction, especially its ability to maintain prediction accuracy comparable to centralized models while ensuring privacy security, remain insufficiently verified on clinically screened, multi-center EHR datasets\cite{b40}. To address this research gap, we propose a federated learning-based distributed collaborative modeling framework for early sepsis prediction across multiple centers, aiming to balance privacy preservation with predictive performance.

To ensure clinical relevance and data quality, we employ strict preprocessing strategies based on clinical guidelines: (1) Rigorous inclusion and exclusion criteria were applied to identify eligible samples. Specifically, patients were included if they were older than 18 years, had an ASA physical status of 2–4, underwent abdominal surgery, and had no preoperative sepsis, infection, or other severe complications; excluding patients who received local anesthesia, superficial surgery, those with repeated or incomplete surgical records, and samples with missing values exceeding 30\%. (2) standardized data normalization and missing value imputation to reduce data heterogeneity and bias across the three participating hospitals, ensuring the consistency and reliability of experimental data\cite{b49,b43,b6,b48,b7}.

\section{Related Work}
\subsection{Traditional Sepsis Prediction Methods}
Early sepsis prediction mainly relies on machine learning models trained on centralized datasets. Random Forest, XGBoost, and SVM have been widely used for clinical risk assessment due to their interpretability and efficiency on small-scale curated data \cite{b30,b32}. However, these methods face two critical limitations in multi-center scenarios: first, they heavily depend on manual feature engineering, which fails to capture complex hidden patterns in high-dimensional electronic health records (EHRs) \cite{b29}; second, centralized training requires aggregating raw medical data, leading to severe privacy leakage risks under strict regulatory constraints \cite{b31}.
\subsection{Federated Learning for Multi-Center Medical Data}
Federated learning (FL) has become a promising technical framework for multi-center medical collaboration, as it enables institutions to jointly train models by aggregating model parameters rather than sharing raw data.\cite{b16,b17,b18,b19}. This approach enables cross-institutional collaborative model training without raw medical data leaving local hospitals, which not only avoids privacy risks associated with data sharing but also integrates multi-center data to enhance model generalization\cite{b20,b21,b22}.

In the context of sepsis prediction, federated learning has been validated to achieve performance comparable to centralized training while refraining from sharing raw electronic health records (EHRs)\cite{b29}. However, existing relevant research still exhibits significant limitations: first, most studies conduct validation using simulated datasets or single-center expanded datasets, lacking support from multi-center real clinical data screened under strict inclusion criteria, which undermines the clinical applicability of research conclusions; second, data heterogeneity across medical institutions leads to performance degradation of federated models, yet current research offers simplistic solutions that fail to fully account for the specificity of clinical data\cite{b23,b24,b25}.
\subsection{Differential Privacy Enhanced Federated Learning}
To mitigate privacy risks in federated learning, differential privacy (DP) mechanisms have been increasingly incorporated into FL frameworks. The core idea is to inject noise during the model parameter update phase at local clients to satisfy $\epsilon$-differential privacy requirements and mitigate privacy security risks arising from parameter leakage\cite{b26,b27,b28}.

Existing federated learning methods combined with differential privacy still face two core challenges in sepsis prediction scenarios: first, static noise injection strategies cannot adapt to the dynamic changes in parameter sensitivity during model training, easily resulting in either excessive privacy protection leading to significant model performance loss or insufficient noise injection causing privacy defense failure; second, privacy security verification is limited in scope—most studies only prove privacy protection effects through theoretical derivation, lacking validation of resistance against advanced attack methods such as model inversion and data reconstruction on real clinical data\cite{b37,b38,b39,b40,b41,b42}. This makes it difficult to assess the actual privacy protection strength of the proposed frameworks.

\section{Methodology}

\subsection{Data Preprocessing}

All data were processed at the individual patient level to maintain the integrity of clinical records and avoid information leakage between the training and test sets. Eligible clinical cases were obtained from three Grade A tertiary hospitals in China, including Southwest Hospital of Third Military Medical University, Xuanwu Hospital of Capital Medical University, and West China Hospital of Sichuan University, between May 2014 and January 2020. Following rigorous screening based on predefined inclusion and exclusion criteria, 648 eligible samples were ultimately retained and partitioned into three client datasets corresponding to the three participating hospitals for subsequent federated learning experiments. The detailed distribution of sepsis samples among the three clients is shown in Table \ref{tab1}, including the number of total samples, sepsis-positive samples, and sepsis-negative samples in the training and test sets of each client.In this study, 27 clinical features were selected as inputs for the prediction model, covering continuous features, binary features, and categorical features\cite{b43}.To ensure the consistency of data representation across multi-center datasets, a ColumnTransformer was adopted to perform heterogeneous preprocessing for different feature types:

\begin{itemize} 
    \item \textbf{Continuous features}: Standardized using StandardScaler to eliminate the influence of dimensional differences; 
    \item \textbf{Binary features}: Retained in their original form without additional transformation; 
    \item \textbf{Categorical features}: Encoded using OneHotEncoder for one-hot encoding.
\end{itemize}

\begin{table}[htbp]
\caption{Distribution of Sepsis Data Among Multi-center Hospitals}
\centering
\small % 改用small字号（比scriptsize大，IEEE常用可读字号）
\setlength{\tabcolsep}{9pt} % 列间距调至4pt（比2pt宽松，比默认6pt紧凑）
\begin{tabular}{ccccc} 
\toprule
\textbf{Dataset} & \textbf{Split} & \textbf{Total} & \textbf{Sepsis}& \textbf{non-Sepsis}\\
\midrule
\multirow{2}{*}{Hospital A} & Training & 371 & 123 & 248 \\
                            & Testing  & 167 & 50  & 117 \\
\midrule
\multirow{2}{*}{Hospital B} & Training & 58  & 8   & 50  \\
                            & Testing  & 28  & 7   & 21  \\
\midrule
\multirow{2}{*}{Hospital C} & Training & 24  & 24  & 0   \\
                            & Testing  & —   & —   & —   \\
\midrule
\multirow{2}{*}{\textbf{Total}} & \textbf{Training} & \textbf{453} & \textbf{155} & \textbf{298} \\
                                & \textbf{Testing}  & \textbf{195} & \textbf{57}  & \textbf{138} \\
\bottomrule
\end{tabular}
\label{tab1}
\end{table}

 \subsection{Federated Learning Framework}
The classical FedAvg (Federated Averaging) algorithm was employed to achieve distributed collaborative training of the end-to-end Autoencoder–MLP model using multi-center clinical data. This strategy enables the global model to be jointly optimized while ensuring that raw clinical data remain within each hospital, thereby safeguarding the privacy of both medical institutions and patients\cite{b40}. The federated learning framework is composed of one central server and three local clients. The central server is responsible only for initializing the global model and aggregating parameters uploaded by the local models, without accessing any raw clinical data. The detailed federated training procedure is illustrated in Fig. \ref{fig1}:

% 第二步：修改figure环境
\begin{figure}[H] % 用[H]替代默认的[htbp]，强制固定位置
    \centering
    \includegraphics[width=1\linewidth]{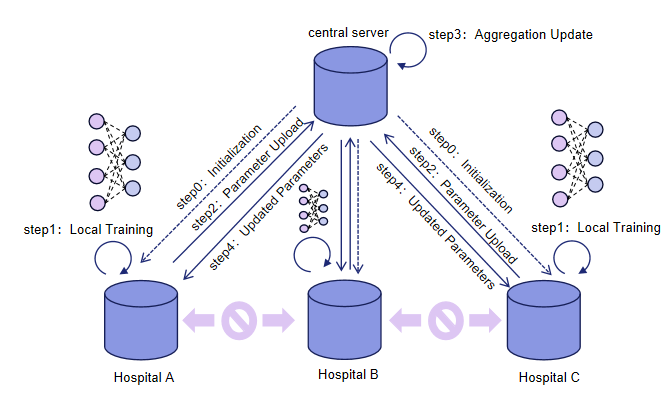}
    \caption{Federated Learning Framework for Sepsis Prediction}
    \label{fig1}
\end{figure}

Step0: \textbf{Initialization} – The central server first initializes the global model parameters $ \theta $ and then distributes them to each participating client.

Step1: \textbf{Local Training} – Each client $i$ performs model optimization using its own local dataset and generates the updated local parameters $ \theta_i $.

Step2: \textbf{Parameter Upload} – After local training, each client uploads its updated model parameters $ \theta_i $ to the central server.

Step3: \textbf{Aggregation Update} – The central server updates the global parameters $ \theta $ by performing weighted averaging according to the sample size $ n_i $ of each client:
\begin{equation}
\theta=\sum_{i=1}^{N}\frac{n_i}{n}\theta_i\label{1}
\end{equation}
where $ N $ is the number of participants, and $ n $ is the total data volume. 

Step4: \textbf{Updated Parameter Distribution} – The central server redistributes the aggregated global parameters $ \theta $ to each client, and the federated training process then proceeds to the next communication round.

 \subsection{Privacy Security Analysis}
Privacy protection capability of federated learning for multi-center sepsis prediction was evaluated by comparing the privacy security characteristics of centralized training and federated learning paradigms from two core dimensions: data access rights and sources of privacy leakage risk. The analysis focused on two key points: whether raw clinical data leaves local medical institutions, and the potential carrier of privacy leakage risk in each training paradigm. 

In the centralized training paradigm, the critical privacy vulnerability derives from the full centralization of raw data. To accomplish model training, all raw clinical data containing sensitive patient information must be transmitted from the three participating hospitals to a central server for unified storage. The server gains direct access to all patients' raw privacy data in this process, making data reconstruction unnecessary for any unauthorized access. Privacy leakage risk in this paradigm stems directly from the exposure of raw clinical data instead of model parameter transmission, and a data breach or malicious attack on the central server would directly compromise the core clinical privacy of all multi-center patients. 

The federated learning (FedAvg) paradigm fundamentally reshapes the data flow mode and thus eliminates the inherent risk of centralized raw data exposure. In the federated learning framework employed in this study, each hospital acts as an independent local client and retains exclusive control over its proprietary raw clinical data. Only locally updated model parameters are uploaded to the central server for aggregation, whereas all raw clinical data are retained within each medical institution and never leave the local environment. 

To assess the resistance of federated learning to such potential privacy risks, a data reconstruction attack simulation experiment was designed. In this experiment, we assumed that a malicious attacker had compromised the central server and sought to reconstruct the original patient-level clinical data from the model parameters exchanged during federated training\cite{b42}. 

The effectiveness of the data reconstruction attack was evaluated by quantifying the discrepancy between the reconstructed clinical label probabilities and the true patient sepsis labels. The Encoder-MLP model adopted for sepsis prediction in this study is defined as:

\begin{equation}
f_\theta(X)=\sigma(W_2\cdot \text{Encoder}(W_1X+b_1)+b_2)\label{2}
\end{equation}
where $X\in \mathbb{R}^{N\times D}$ denotes the clinical feature matrix containing $N$ samples and $D$ features, $\theta=\lbrace W_1,b_1,W_2,b_2 \rbrace$ represents the weight and bias parameters of the MLP model, $\text{Encoder}(\cdot)$ denotes the feature encoding layer, and $\sigma(\cdot)$ is the Sigmoid activation function, which outputs the predicted sepsis probability ranging from 0 to 1.

The MAE and RMSE metrics for quantifying the privacy leakage risk induced by data reconstruction attacks are defined as:

\begin{equation}
\text{MAE} = \frac{1}{N} \sum_{i=1}^N \left| y_i - \tilde{y}_{\text{pred},i} \right|\label{4}
\end{equation}

\begin{equation}
\text{RMSE} = \sqrt{\frac{1}{N} \sum_{i=1}^N \left( y_i - \tilde{y}_{\text{pred},i} \right)^2}\label{5}
\end{equation}
where $\tilde{y}_{\text{pred},i}$ denotes the sepsis probability of the $i$-th patient reconstructed by the attacker from the leaked model parameters transmitted during federated training.

Building on the above privacy inference analysis based on DLG, to further provide quantifiable privacy protection guarantees at the mechanism level, this paper adopts a federated learning differential privacy framework with local parameter update perturbation, whose core lies in injecting Laplace noise during the client-side model parameter update phase\cite{b41}. 

After completing local training on its own dataset, each client computes the parameter difference between the updated local model and the current global model:

\begin{equation}
\Delta \theta_i^t = \theta_i^t - \theta_{\text{global}}^{t-1}\label{6}
\end{equation}

where $\theta_i^t$ denotes the local model parameters of the $i$-th client in the $t$-th training round, and $\theta_{\text{global}}^{t-1}$ represents the global model parameters from the $(t-1)$-th round.

Sensitivity is a key parameter in differential privacy, as it quantifies the maximum influence that a single data record can have on the query output. In this study, a data-driven dynamic sensitivity estimation method was adopted:

\begin{equation}
\Delta f = \frac{\|\Delta \theta_i^t\|_1}{\sqrt{N_i}}\label{7}
\end{equation}

where $\|\cdot\|_1$ denotes the L1 norm, and $N_i$ represents the number of samples held by the $i$-th client.

Based on the Laplace mechanism, noise is added to the parameter updates to ensure $\epsilon$-differential privacy:

\begin{equation}
\tilde{\Delta \theta}_i^t = \Delta \theta_i^t + \text{Lap}\left(0, \frac{\Delta f}{\epsilon}\right)\label{8}
\end{equation}

where $\text{Lap}\left(0, \beta\right)$ denotes the Laplace distribution with mean 0 and scale parameter $\beta$, while $\epsilon$ represents the privacy budget.

The server collects the noisy parameter updates from all clients and aggregates them weighted by the sample size to obtain the new global model:

\begin{equation}
\theta_{\text{global}}^t = \theta_{\text{global}}^{t-1} + \sum_{i=1}^K \frac{N_i}{N_{\text{total}}} \cdot \tilde{\Delta \theta}_i^t\label{8}
\end{equation}
where $K$ denotes the number of clients.

% 需确保文档开头加载必备包：
% \usepackage{booktabs}  % 三线表核心包（必须）
% \usepackage{amsmath}   % 可选，兼容数学符号

\begin{table*}[htbp] % *表示跨双栏
\caption{Performance Comparison of Baseline, Centralized, and Federated Models}
\centering
\small % 跨栏可用\small，无需缩到\scriptsize
\setlength{\tabcolsep}{8pt} % 调整列间距，适配三线表无竖线的视觉效果
\begin{tabular}{lccccccc} % 1个左对齐列 + 7个数值列（无竖线，三线表规范）
\toprule % 顶线（三线表第一行）
\textbf{Subgroup}       & \textbf{ROC AUC} & \textbf{Sensitivity} & \textbf{Specificity} & \textbf{PPV}  & \textbf{NPV}  & \textbf{F1 Score} & \textbf{Accuracy} \\
\midrule % 中线（三线表第二行）
Baseline                & 0.8789           & 0.5789               & 0.9492               & 0.8250        & 0.8451        & 0.6804            & 0.8410            \\
Central                 & 0.9027           & 0.6315               & 0.9492               & 0.8372        & 0.8518        & 0.7200            & 0.8564            \\
FedAvg                  & 0.8902           & 0.5614               & 0.9275               & 0.7619& 0.8366& 0.6464            & 0.8205            \\
\bottomrule % 底线（三线表第三行）
\end{tabular}
\label{tab2}
\end{table*}

\begin{table*}[htbp]  % *号表示跨两栏（双栏文档核心）
\caption{Predictive Performance of Federated Learning with Differential Privacy}
\centering
\small
\setlength{\tabcolsep}{8pt}  % 调整列间距，适配跨栏宽度
\begin{tabular}{lccccccc}  % 8列：1文本列+7指标列（修正原列数错误）
\toprule  % 顶线（三线表第一行）
\textbf{Subgroup} & \textbf{ROC AUC} & \textbf{Sensitivity} & \textbf{Specificity} & \textbf{PPV} & \textbf{NPV} & \textbf{F1 Score} & \textbf{Accuracy} \\
\midrule  % 中线（三线表第二行）
$\epsilon$=1  & 0.6375 & 0.6315 & 0.5942 & 0.3913 & 0.7961 & 0.4832 & 0.6051 \\
$\epsilon$=5  & 0.6823 & 0.7017 & 0.6594 & 0.4597 & 0.8425 & 0.5555 & 0.6717 \\
$\epsilon$=10 & 0.7922 & 0.7719& 0.6594& 0.4835& 0.8750& 0.5945& 0.6923\\
No-DP        & 0.8902 & 0.5614 & 0.9275 & 0.7619& 0.8366& 0.6464 & 0.8205 \\
\bottomrule  % 底线（三线表第三行）
\end{tabular}
\label{tab3}
\end{table*}

\section{Results}
To evaluate the predictive performance and privacy-preserving benefits of the proposed federated learning framework, we conducted a comprehensive comparison with two baseline models: a fixed ensemble learning approach (KNN+SVM) and a centralized end-to-end autoencoder--MLP model. The evaluation was performed using the multi-center clinical dataset collected from three tertiary hospitals and focused on both prediction performance and privacy protection capability.

As shown in Table \ref{tab2}, the centralized end-to-end model (Central) achieved an ROC AUC of 0.9027, sensitivity of 0.6315, specificity of 0.9492, PPV of 0.8372, NPV of 0.8618, an F1 score of 0.72, and an accuracy of 0.8564, demonstrating superior performance over the fixed ensemble baseline across all major evaluation metrics. In comparison, the federated learning model (FedAvg) yielded similarly competitive results, with an ROC AUC of 0.8902, sensitivity of 0.5614, specificity of 0.9275, PPV of 0.7619, NPV of 0.8366, an F1 score of 0.6464, and an accuracy of 0.8205. These results suggest that the proposed federated framework achieves predictive performance close to that of the centralized end-to-end model, with only limited performance loss, while eliminating the need to directly exchange raw clinical data.

To evaluate the privacy-preserving capability of the federated learning framework, we simulated a data reconstruction attack in which the central server was assumed to be compromised by a malicious attacker. The attacker attempted to infer the original patient-level clinical data from the model parameters exchanged during federated training. As shown in Table \ref{tab3}, under the No-DP FedAvg framework, the reconstruction error was measured using mean absolute error (MAE) and root mean square error (RMSE), which were 0.2264 and 0.4018, respectively. These error values indicate a substantial discrepancy between the reconstructed data and the original ground-truth data. Specifically, the reconstructed binary features failed to clearly differentiate patients' clinical attributes, while the reconstructed continuous features also showed noticeable deviations from their true values, making them unsuitable for clinical decision-making.

\begin{table}[htbp]
\caption{Data Reconstruction Attack Error Under Federated Learning Framework}
\centering
\small % 缩小字体适配页面（可选）
% 加载booktabs包后使用三线表（需在文档开头添加 \usepackage{booktabs}）
\setlength{\tabcolsep}{27pt}
\begin{tabular}{lcc}
\toprule % 顶线
\textbf{Subgroup}       & \textbf{MAE}     & \textbf{RMSE}        \\
\midrule % 中线

$\epsilon$=1& 0.6288       & 0.7889               \\
$\epsilon$=5& 0.4906       & 0.6646               \\
$\epsilon$=10& 0.4485       & 0.4919               \\
No-DP   & 0.2264           & 0.4018               \\
\bottomrule % 底线
\end{tabular}
\label{tab4}
\end{table}

Building on this, we further introduced a Laplace differential privacy mechanism to enhance privacy protection by injecting noise into client-side parameter updates. As shown in Table \ref{tab3} and Table \ref{tab4}, as the privacy budget $\epsilon$ decreases, indicating stronger privacy protection, the privacy inference error increases significantly, while the predictive performance degrades to varying degrees, presenting a clear privacy–performance trade-off.

\section{Conclusion}
This study focuses on sepsis prediction in real-world multi-center clinical scenarios, aiming to evaluate the feasibility, privacy-preserving capability, and predictive stability of the FedAvg-based federated learning framework using real clinical data. The primary objective is to reconcile the trade-off between privacy preservation and robust model performance in multi-center medical collaboration.

In real-world clinical data scenarios, a major bottleneck in multi-center medical collaboration is that raw clinical data contain sensitive patient information, making cross-institutional data sharing subject to strict regulatory constraints and potential privacy leakage risks. By applying the federated learning framework to real clinical datasets from three tertiary hospitals, this study fundamentally changes the data flow pattern of traditional centralized training.

Experimental verification results demonstrate that the federated learning framework exhibits significant application value in real-world medical data environments: First, it achieves excellent predictive performance. Compared with traditional centralized training models, its core metrics such as ROC AUC, sensitivity, specificity, and accuracy remain highly consistent, with only acceptable slight performance loss, which fully meets the requirements of clinical practical applications for predictive accuracy. Second, it provides reliable privacy protection. Through data reconstruction attack simulation experiments, it is confirmed that malicious attackers cannot effectively recover patients' real clinical information from the transmitted model parameters. The reconstruction errors are significant and have no clinical decision-making significance, fully verifying the privacy protection capability of the framework in real medical data scenarios. 

the application of the federated learning framework in real multi-center medical data has successfully achieved the core goal of collaborative modeling without data sharing. It not only guarantees the model performance of multi-center medical collaboration but also strictly complies with the regulatory requirements for medical data privacy protection, providing a practical and feasible implementation solution to address the industry pain point of cross-institutional medical data collaboration.

\end{document}